\documentclass{article}
\usepackage{ijcai21}

\usepackage{times}
\usepackage{soul}
\usepackage{url}
\usepackage{hhline}
\usepackage[hidelinks]{hyperref}
\usepackage[utf8]{inputenc}
\usepackage[small]{caption}
\usepackage{graphicx}
\usepackage{amsmath}
\usepackage{amssymb}
\usepackage{amsthm}
\usepackage{multirow}
\usepackage{amsfonts}
\usepackage{booktabs}
\usepackage{algorithm}
\usepackage{diagbox}
\usepackage{subfigure}
\usepackage{algorithmic}
\usepackage{enumitem}

\usepackage{listofitems}
\newcommand\textlist[3][\cr]{%
  \readlist\indices{#3}%
  \foreachitem\x\in\indices{%
    \ifnum\xcnt=1\else, \fi$#2_{\x}$%
  }%
  \ifx\cr#1\relax, \ldots\else%
    \if\relax#1\relax\else, \ldots, $#2_{#1}$\fi%
  \fi%
}
\title{Smart Contract Vulnerability Detection: From Pure Neural Network to Interpretable Graph Feature and Expert Pattern Fusion}

\author{
Zhenguang Liu$^1$\footnotemark[1] \and
Peng Qian$^1$\footnotemark[1] \and
Xiang Wang$^2$\and
Lei Zhu$^3$\and
Qinming He$^1$\And
Shouling Ji$^1$\footnotemark[2]
\affiliations
$^1$Zhejiang University\\
$^2$National University of Singapore\\
$^3$Shandong Normal University\\
\emails
\{liuzhenguang2008, messi.qp711, xiangwang1223, leizhu0608\}@gmail.com,
\{hqm, sji\}@zju.edu.cn
}

\hypersetup{draft}

\begin{document}
\maketitle

\renewcommand{\thefootnote}{\fnsymbol{footnote}} 
\footnotetext[1]{The first two authors have equal contribution.}
\footnotetext[2]{Corresponding author.}

\begin{abstract}
Smart contracts hold digital coins worth billions of dollars, their security issues have drawn extensive attention in the past years. Towards smart contract vulnerability detection, conventional methods heavily rely on  fixed expert rules, leading to low \emph{accuracy} and poor \emph{scalability}. Recent deep learning approaches alleviate this issue but fail to encode useful expert knowledge. In this paper, we explore combining  deep learning with expert patterns  in an explainable fashion. Specifically, we develop automatic tools to extract expert patterns from the source code. We then cast the code into a semantic graph to extract deep graph features. Thereafter, the \emph{global graph feature} and \emph{local expert patterns} are fused to cooperate and approach the final prediction, while yielding their interpretable weights. Experiments are conducted on all available smart contracts with source code in two platforms, \emph{Ethereum} and \emph{VNT Chain}. Empirically, our system significantly outperforms state-of-the-art methods. Our code is released.
\end{abstract}

\section{Introduction}
Blockchain has attracted extensive attention in the past few years. The worldwide miners (bookkeeping nodes) obey a consensus protocol to maintain a secure and shared transaction ledger, which is termed a blockchain \cite{hewa2020survey}. In the blockchain network, the consensus protocol enforces the transactions immutable once recorded in the distributedly copied ledger, endowing the blockchain with \emph{tamper-resistant} and \emph{decentralization} nature.

Smart contracts are programs running on top of a blockchain system \cite{wang2019blockchain,oyente}. A smart contract can be specially designed by developers to implement arbitrary rules for managing digital assets. Attributing to the immutable nature of blockchain, a smart contract cannot be updated once deployed. Thus, the defined rules of a smart contract are formulated as program code and are automatically executed, which is impartial for all parties that interact with the contract. Smart contracts make the automatic execution of contract terms possible, facilitating complex decentralized applications (DApps) \cite{dapps}. 

So far, millions of smart contracts have been deployed on various blockchain platforms, controlling digital currency worth more than 10 billion dollars. Holding so much wealth, however, makes smart contracts attractive enough to malicious attackers. In 2016, attackers exploited the reentrancy vulnerability of \emph{The DAO} contract\footnotemark[3] to steal Ether (i.e., Cryptocurrency of Ethereum) worth 60 million dollars. This case is not isolated and several security vulnerabilities of smart contracts are disclosed every year \cite{Multisig}. 

\footnotetext[3]{The DAO contract, 2016. \url{http://etherscan.io/address/0xbb9bc244d798123fde783fcc1c72d3bb8c189413}}

\begin{figure*}
\centering
\includegraphics[width=17.8cm]{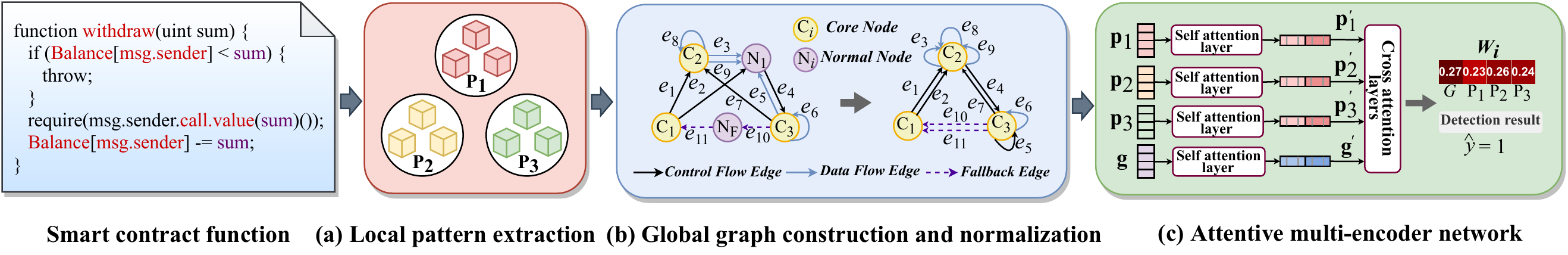}
\caption{The overall architecture of our proposed method. (a) The local expert pattern extraction tool for extracting vulnerability-specific expert patterns. (b) The graph construction and normalization module for transforming the code into a global semantic graph. (c) The attentive multi-encoder network, which combines expert patterns and the graph feature for vulnerability detection and outputting explainable weights.}
\label{fig_overview}
\end{figure*}

The main issues that may easily lead to smart contract vulnerabilities are twofold. \emph{First}, the programming languages and tools are still new and crude, which leaves plenty of rooms for  misunderstandings in the built-in functions and tools \cite{oyente}. \emph{Second}, due to the immutable nature of smart contracts, developers are required to anticipate all possible states (\emph{e.g., stack status}) and environments that the code may encounter in the future, which is obviously difficult. 

Existing methods on smart contract vulnerability detection  can be roughly cast into two categories. The first line of work \cite{oyente,tsankov2018securify,contractfuzzer} utilized classical \emph{static analysis} and \emph{dynamic execution} techniques to identify vulnerabilities. Unfortunately, they fundamentally rely on several fixed expert rules, while the manually defined patterns bear the inherent risk of being \emph{error-prone} and some complex patterns are \emph{non-trivial} to be covered. Meanwhile, crafty attackers may easily bypass the fixed patterns using small tricks. Another line of work \cite{Wesley,ijcai20} explored using deep learning models to deal with complex contract data, having achieved much improved accuracy. Due to the \emph{black-box} nature, however, they fail to encode useful expert knowledge and mostly have poor interpretability. This motivates us to consider whether we could combine neural networks with classical expert patterns, where neural networks contribute their ability to handle the complex code semantic graph while  expert patterns contribute precise and valuable local information. \textbf{More importantly}, we seek an explainable solution which could tell the weights of different features.

\paragraph{Our method.} In this paper, we propose a new system beyond pure neural networks that can automatically detect vulnerabilities and incorporate expert patterns into networks in an explainable fashion. {In particular}, (1) we develop automatic tools to extract  \emph{vulnerability-specific}  {expert patterns}. (2) Then, we exploit a graph structure to frame the rich control-flow and data-flow semantics of the function code. Upon the graph, a graph neural network is employed to extract the deep graph feature. (3) Finally, we propose an attentive multi-encoder network to interpretably  fuse the \emph{global} graph feature and \emph{local} expert patterns. Extensive experiments are conducted on all the 40k contracts in two benchmark datasets, demonstrating significant improvements over state-of-the-art: \emph{accuracy} from $84\%$ to $90\%$, $83\%$ to $87\%$, $75\%$ to $80\%$ on three types of vulnerabilities respectively. More importantly, our model is able to explain its label prediction, give warnings of high weighted local patterns, and provide a grand picture of the significance of different features. 
 
\paragraph{Contributions.} The key contributions of this work are: 1) We investigate combining vulnerability-specific expert patterns with neural networks in an explainable way. To the best of our knowledge, we are the first to prob the combination interpretably. 2) In the method, we 
present a simple but effective multi-encoder network for feature fusion. 3) Our method sets the new state-of-the-art and provides novel insights. To facilitate future research, our implementations are  released at  \url{https://github.com/Messi-Q/AMEVulDetector}. We would like to point out that different from \cite{tkde2021}, this work focuses mainly on the explainability of the expert pattern and deep graph feature combination, and  offers a grand picture on the  importance of different features.

\section{Problem}
Given the source code of a smart contract, we seek to develop a fully automatic approach that can detect vulnerability at fine-grained \emph{function level} and is \emph{interpretable}. In other words, presented with a smart contract function $f$, we are to predict its label $\hat{l}$ and output the associated weights $\{w_i\}_{i=1}^n$ simultaneously, where $\hat{l}$ = 1 represents $f$ has a vulnerability of a certain type and $\hat{l} = 0$ denotes $f$ is safe. Weight $w_i$ explains the importance of the $i^{th}$ feature in predicting the label. In this work, we concentrate on three common vulnerabilities:

(1) Reentrancy is a well-known vulnerability that has brought about the notorious DAO attack. When a smart contract function $f_1$ transfers money to a recipient contract $C$, the fallback function  $f_2$  of $C$ will be automatically triggered. Function $f_2$ may invoke $f_1$ for conducting an invalid second-time money transfer. Since the current execution of $f_1$ waits for the first-time transfer to finish, the balance of $C$ is not  reduced yet, $f_1$ thus may wrongly believe that $C$ still has enough balance and transfers money to $C$ again. More specifically, the expected execution trace is $f_1 \xrightarrow{{transfer}} C \xrightarrow{trigger} f_2 \rightarrow end$, 
whereas the actual trace is ${f_1} \xrightarrow{{transfer}} C \xrightarrow{trigger} f_2 \xrightarrow{invoke} f_1  \xrightarrow{{transfer}} C \xrightarrow{trigger} f_2 \rightarrow end$.  Exploiting the reentrancy vulnerability, an attacker may succeed in obtaining 20 Ether although his/her balance is 10 Ether. Note that attackers can steal more money by invoking $f_1$ more than one time, which can be exemplified as ${f_1} \xrightarrow{{transfer}} C \xrightarrow{trigger} f_2 \xrightarrow{invoke} f_1 \xrightarrow{{transfer}} C \xrightarrow{trigger} f_2   \xrightarrow{invoke} f_1 \xrightarrow{{transfer}} C \xrightarrow{trigger} f_2 \rightarrow end$.

(2) Block timestamp dependence happens when a function uses block timestamp as a condition to perform critical operations, e.g., using \emph{block.timestamp} of a future block as the source to generate random numbers so as to determine the winner of a game. The miner who mines the block has the freedom to set the timestamp of the block as long as it is within a short time interval \cite{contractfuzzer}. Thus, miners may manipulate block timestamp to gain illegal benefits.

(3) Infinite loop is conventionally considered as a loop bug which unintentionally iterates forever, e.g., a \emph{for} loop with no exit condition. Distinct from conventional programs, users have to pay a fee for executing each line of smart contract code. The fee is approximately proportional to how much code needs to run. For a function with infinite loop, its execution will run out of gas and be aborted. In such a case, the execution consumes a lot of gas but all the gas is consumed in vain since  the execution is unable to change any state.

\section{Method}
The overview of the proposed system is illustrated in Fig.~\ref{fig_overview}, which consists of three components: 1) a local expert pattern extraction tool, which extracts expert patterns of a specific vulnerability from the function code; 2) a graph construction and normalization module, which transforms the function code into a code semantic graph; and 3) an attentive multi-encoder network that combines local expert patterns and the global graph feature for vulnerability detection and outputs explainable weights. In what follows, we introduce the three components one by one.

\subsection{Local Expert Pattern Extraction}
\label{pattern_extraction}
Following \cite{tkde2021}, we design corresponding expert patterns for three types of vulnerabilities respectively. Then, we implement a fully automatic tool to extract  expert patterns from the function code. Specifically, the patterns for different vulnerabilities are defined as:

\paragraph{Reentrancy.} Technically, the reentrancy vulnerability occurs when a \emph{call.value} invocation (i.e., a built-in money transfer function) can call back to itself through a chain of calls. That is, \emph{call.value} is successfully re-entered to perform unexpected repeat money transfers. For the reentrancy vulnerability, we design three local patterns. \textbf{(1)} \textbf{enoughBalance} concerns whether there is a check on the sufficiency of the user balance before transferring to a user. \textbf{(2)} \textbf{callValueInvocation} models whether there exists  an invocation to \emph{call.value} in the function. \textbf{(3)} \textbf{balanceDeduction} checks whether the user balance is deducted \emph{after} money transfer, which considers the fact that the money stealing can be avoided if the user balance is deducted each time \emph{before} money transfer. 

\paragraph{Block timestamp dependence.} Generally, the timestamp dependency vulnerability exists when a smart contract is conventionally considered as using \emph{block.timestamp} as part of the conditions to perform critical operations \cite{contractfuzzer}. We design three local patterns for the timestamp dependence vulnerability. \textbf{(1)} \textbf{timestampInvocation} models whether there exists  an invocation to opcode \emph{block.timestamp} in the function. \textbf{(2)} \textbf{timestampAssign} checks whether the value of \emph{block.timestamp} is assigned to other variables or passed to a function as a parameter, namely whether \emph{block.timestamp} is actually used. \textbf{(3)} \textbf{timestampContaminate} verifies if \emph{block.timestamp} may contaminate the triggering condition of a critical operation (e.g., money transfer).

\paragraph{Infinite loop.} Specifically, we define three local patterns for the infinite loop vulnerability as follows. \textbf{(1)} \textbf{loopStatement} checks whether the function exists a loop statement such as \emph{for} and \emph{while}. \textbf{(2)} \textbf{loopCondition} validates whether the exit condition can be reached. For example, for a \emph{while} loop, its exit condition $i<9$ cannot be reached if $i$ is never updated in the loop. \textbf{(3)} \textbf{selfInvocation} concerns whether the function invokes  itself and the self-invocation is not in an \emph{if} statement. This considers the fact that if the self-invocation statement is not in an \emph{if} statement, the self-invocation loop will never terminate.

\paragraph{Implementations.}  Our open-sourced  tool to extract the designed local expert patterns is released on Github. These patterns are consistent with those in our previous work \cite{tkde2021}. As introduced in  \cite{tkde2021}, the pattern extraction tool scans the source code of a function several times. Particularly, simple patterns such as \emph{callValueInvocation} and \emph{timestampInvocation} can be directly extracted by keyword matching. Patterns such as \emph{enoughBalance}, \emph{balanceDeduction}, \emph{loopStatement}, \emph{timestampAssign}, \emph{loopCondition}, and \emph{selfInvocation} are obtained by syntactic and semantic analysis. Complex pattern \emph{timestampContaminate} is extracted by taint analysis, where we follow the traces of the data flow and flag all the variables that may be affected along the traces.

\begin{table}
 \renewcommand\arraystretch{1.1}
 \renewcommand{\multirowsetup}{\centering}
 \centering
 \resizebox{0.48\textwidth}{!}{
      \begin{tabular}{|c|c|}
      \hline
       \textbf{Vulnerabilities} & \textbf{Core Nodes} \\
       \hline
       \multirow{3}{*}{Reentrancy} & call.value invocation  \\
        & a function that contains call.value  \\
        & \textbf{the variable: correspond to user balance}  \\
        \hline
       \multirow{3}{*}{Timestamp dependence} & block.timestamp invocation \\
        &  block.number invocation  \\
         & \textbf{a variable: affect critical operation}  \\
         \hline
       \multirow{3}{*}{Infinite loop} & for \\
         & while  \\
         & \textbf{self-call\ function}  \\
       \hline
     \end{tabular}}      
  \caption{Core nodes are modeled for the three types of vulnerabilities. Differences with [Zhuang \emph{et al.}, 2020] are highlighted in bold.}
 \label{core_nodes}
\end{table}

\begin{table}
 \renewcommand\arraystretch{1.1}
  \renewcommand{\multirowsetup}{\centering}
\centering
\resizebox{0.48\textwidth}{!}{
\begin{tabular}{|c|c|}
\hline
\textbf{Semantic Edge} & \textbf{Type} \\
\hline
assert\{ X \} & \multirow{10}{*}{\begin{tabular}[c]{@{}c@{}}Control-flow edges\end{tabular}}  \\
require\{X\} &  \\
if\{X\} & \\
if\{...\} else \{X\} & \\
if\{...\} revert & \\
if\{...\} throw & \\
if\{...\} then \{X\} & \\
while\( \{ X\}\) do\{...\} & \\
for\( \{ X\}\) do\{...\} & \\
natural sequential relationships & \\
\hline
assign\{X\} & \multirow{2}{*}{\begin{tabular}[c]{@{}c@{}}Data-flow edges\end{tabular}} \\
access\{X\} & \\
\hline
from call.value node to fallback function & \multirow{2}{*}{\begin{tabular}[c]{@{}c@{}}Fallback edges\end{tabular}} \\
from fallback node to function under test & \\
\hline
\end{tabular}}
 \caption{Key semantic edges, which fall into three categories.}
 \label{semantic_edges}
\end{table}

\begin{figure*}
\centering
\includegraphics[width=17.8cm]{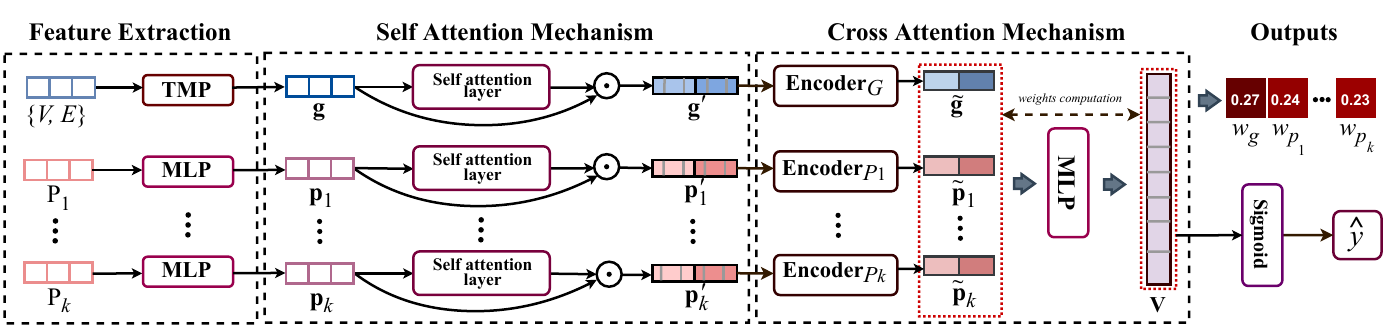}
\caption{The attentive multi-encoder network, consisting of a self-attention mechanism and a cross-attention mechanism. It combines local pattern features and the global graph feature for vulnerability detection, and outputs interpretable weights for all features.}
\label{fig_network}
\end{figure*}

\subsection{Graph Construction and Normalization}
\label{graph_feature}
One may directly use combinations of aforementioned patterns to predict whether the function has a certain vulnerability. However, these fixed patterns are shown to have difficulties in handling relatively complex attacks and are trivial to be bypassed by adversaries \cite{peng}. Therefore, we further propose  to model the control flow and data flow of the entire source code into a semantic graph, and adopt graph neural networks to handle it. Thereafter, the extracted global graph feature and the local expert patterns could supplement each other towards a more precise and explainable label prediction. 

\paragraph{Graph construction.} Different program elements in a function are not of equal importance in detecting vulnerabilities. Therefore, we extract two categories of nodes, {i.e.}, \emph{core nodes} and \emph{normal nodes}. (1) Core nodes symbolize  key invocations and variables in the function code, which are critical for detecting a specific vulnerability. For instance, for reentrancy vulnerability, (i) an invocation to \emph{call.value}, (ii) an invocation to a money transfer function that contains a \emph{call.value} invocation, and (iii) the variable that corresponds to \emph{user balance}, are treated as core nodes. We summarize the core nodes for detecting the three vulnerabilities in Table~\ref{core_nodes}. (2) Invocations and variables that are not extracted as core nodes are modeled as normal nodes, which play an auxiliary role in detecting vulnerabilities. We also construct an extra fallback node to stimulate the fallback function of a virtual attack contract. The fallback node can interact with the function under test and is considered as a normal node. It is worth mentioning that distinct from prior works such as \cite{ijcai20}, which merely model key invocations as core nodes, we propose to further extract key variables as core nodes, given that they are undoubtedly important in detecting vulnerabilities. To characterize rich connections between different nodes, we construct three categories of semantic edges, namely \emph{control flow, data flow}, and \emph{fallback} edges. Each edge describes a path that might be traversed through by the function under test, and the temporal number of the edge is set according to its sequential order in the function code. We summarized key semantic edges in Table \ref{semantic_edges}.

\paragraph{Graph normalization.} It is worth mentioning that different functions corresponding to distinct code semantic graphs, bringing difficulties in training a graph neural network. Moreover, current graph neural networks are inherently flat when propagating information, ignoring that different nodes are not of equal importance. Therefore, we propose to normalize the graph following that of \cite{ijcai20} to remove all normal nodes and merge their features to the nearest core nodes. A simplified example for graph construction and normalization is given in Fig.~\ref{fig_overview}(b).

\subsection{Attentive Multi-Encoder Network} 
For the extracted local expert patterns, we adopt the multiple MLPs (multilayer perceptrons) to encode them into feature vectors. For the normalized global code semantic graph, we utilize a temporal-message-propagation (TMP) graph neural network to transform it into a deep graph feature. Then, the expert pattern features and the graph feature are fused using an {a}ttentive {m}ulti-{e}ncoder network to give the overall label prediction $\hat{y}$, which is illustrated in Fig.~\ref{fig_network}.

\paragraph{Local pattern feature extraction.} Each expert pattern formulates an elementary factor closely related to a specific vulnerability. We utilize a one-hot vector to represent each pattern, and append a digit 0/1 to indicate whether the function under test has this pattern. The vectors for all patterns related to a specific vulnerability are fed into multiple MLPs, each takes care of one pattern. The outputs of the MLPs are concatenated and passed into a feed-forward network to compute the label prediction of the function. This network is trained so that each MLP learns to  extract the feature $\textbf{p} \in \mathbb{R}^{d}$ of a pattern.

\paragraph{Global graph feature extraction.}  To extract the graph feature from the normalized code semantic graph, we use a graph neural network that consists of a \emph{message propagation} phase and an \emph{aggregation} phase \cite{ijcai20}. In the message propagation phase, the network passes information along the edges successively by following their sequential orders in the code, while the aggregation phase outputs the global graph feature $\bold{g}  \in \mathbb{R}^{d}$ by aggregating the final states and original states of all nodes. Following \cite{ijcai20}, at each time step $j$, message flows through the $j^{th}$ edge $e_{j}$ and updates the hidden state of the end node of $e_{j}$ by absorbing information from edge $e_{j}$ and the start node of $e_{j}$. After successively traversing all edges, we extract the graph feature by aggregating the hidden states of all nodes. The original hidden state $h^{0}_{i}$ and the final hidden state $h^{T}_{i}$ of each node are informative in the vulnerability detection task. Therefore, we can obtain the final global graph feature $\bold{g}$ by:
\begin{align}
\label{eq:e3}
\bold{h_i} &= h^{0}_{i} \oplus h^{T}_{i} \\
Gate_{i} &= \sigma (M_{1}(relu(b_1 + M_2\bold{h_i})) + b_2) \\
Output_{i} &= \sigma (M_3(relu(b_3 + M_4\bold{h_i})) + b_4) \\
\bold{g} &= FC\big( \sum\nolimits_{i=1}^{N} (Gate_{i} \odot Output_{i})\big)
\end{align}
where $\oplus$ and $\odot$ denote concatenation and element-wise product. Matrix $M_{j}$ and bias vector $b_{j}$ are network parameters, $N$ is the number of nodes, $\sigma$ is the \emph{softmax} activation layer, and $FC$ is a fully connected layer. Motivated readers may also refer to \cite{ijcai20} for more details of the \emph{message propagation} and  \emph{aggregation} phases.

\paragraph{Fusion network.} As shown in Fig.~\ref{fig_network}, the extracted local pattern features $\{\bold{p}_i\}_{i=1}^k$ and the global graph feature $\bold{g}$ go through a self-attention mechanism and a cross-attention mechanism to output the label prediction $\hat{y}$ and the associated weights $\{w_i\}_{i=1}^n$ for all features. Since the local expert pattern features and the global graph feature are heterogeneous, the self-attention mechanism learns to preliminarily balance different features by computing the coefficients for all features. The cross-attention mechanism then fuses different features to predict the label. 

\paragraph{Self attention mechanism.} Formally, the processes of the self attention for $\bold{p}_i$ and $\bold{g}$ are given by:
\begin{align}
\label{eq:e2}
c_{g} &= FC(\bold{g}), \quad\quad  \ \ \  \bold{g}^{'} = c_{g} \odot \bold{g} \\
c_{p_i} &= FC(\bold{p}_i), \quad\quad {\bold{p}_i}^{'} =c_{p_i}   \odot  \bold{p}_i
\end{align}
where  $c_{g}$ and $c_{p_i}$ are coefficient vectors, $\odot$ denotes the element-wise product, $\bold{g}^{'}$ and ${\bold{p}_i}^{'}$ denote the updated features.

\paragraph{Cross attention mechanism.} After the self attention, we can obtain new features $\bold{g}^{'}$ and $\{{\bold{p}_i}^{'}\}_{i=1}^k$. Next, we consider combining them to detect vulnerabilities. As shown in Fig.~\ref{fig_network}, each feature goes through an encoder, mapping  $\bold{g}^{'}$ and $\{{\bold{p}_i}^{'}\}_{i=1}^k$ into  $\tilde{\bold{g}}$ and $\{\tilde{{\bold{p}}}_i\}_{i=1}^k$, respectively. Then, all the mapped features are concatenated and fed into an MLP, generating a final semantic vector $\textbf{v}$.  The label  $\hat{y}$ is then conveniently computed by $\hat{y} = round(sigmoid(\textbf{v}))$. Technically, we believe that the attention weights could highlight the feature importance. The inner product between each feature and $\textbf{v}$, therefore, is employed to compute the interpretable weights for the global graph feature $\bold{g}$ and each local expert pattern feature $\bold{p}_i$. 
\begin{align}
\label{eq:e4}
w_{g} &= \frac{exp(\tilde{\bold{g}} \cdot \textbf{v})}{exp(\tilde{\bold{g}} \cdot \textbf{v})+\sum_{i=1}^{k}exp(\tilde{{\bold{p}}}_i \cdot \textbf{v})} \\ 
w_{p_{i}} &= \frac{exp(\tilde{{\bold{p}}}_i \cdot \textbf{v})}{exp(\tilde{\bold{g}} \cdot \textbf{v})+\sum_{i=1}^{k}exp(\tilde{{\bold{p}}}_i \cdot \textbf{v})}
\end{align}
where $w_{g}$ denotes the weight of $\bold{g}$, $w_{p_{i}}$ is the weight of $\bold{p}_i$, and $\cdot$ represents inner product.

\renewcommand\arraystretch{1.3}
\begin{table*}
\centering
\resizebox{1.0\textwidth}{!}{
\begin{tabular}{|c|cccc|cccc|c|cccc|}
\hline
\multirow{2}{*}{\textbf{Methods}} & \multicolumn{4}{c|}{\textbf{Reentrancy}} & \multicolumn{4}{c|}{\textbf{Timestamp dependence}} & \multirow{2}{*}{\textbf{Methods}} & \multicolumn{4}{c|}{\textbf{Infinite Loop}}\\
\cline{2-5}\cline{6-9}\cline{11-14} & Acc(\%) & Recall(\%) & Precision(\%) & F1(\%) & Acc(\%) & Recall(\%) & Precision(\%) & F1(\%) & & Acc(\%) & Recall(\%) & Precision(\%) & F1(\%) \\
\hline
Smartcheck & 52.97 & 32.08 & 25.00 & 28.10 & 44.32 & 37.25 & 39.16 & 38.18 & Jolt & 42.88 & 23.11 & 38.23 & 28.81 \\
Oyente & 61.62 & 54.71 & 38.16 & 44.96 & 59.45 & 38.44 & 45.16 & 41.53 & PDA & 46.44 & 21.73 & 42.96 & 28.26 \\
Mythril & 60.54 & 71.69 & 39.58 & 51.02 & 61.08 & 41.72 & 50.00 & 45.49 & SMT & 54.04 & 39.23 & 55.69 & 45.98 \\
Securify & 71.89 & 56.60 & 50.85 & 53.57 & -- & -- & -- & -- & Looper & 59.56 & 47.21 & 62.72 & 53.87 \\
Slither & 77.12 & 74.28 & 68.42 & 71.23 & 74.20 & 72.38 & 67.25 & 69.72 & -- & -- & -- & -- & -- \\
\hline
Vanilla-RNN & 49.64 & 58.78 & 49.82 & 50.71 & 49.77 & 44.59 & 51.91 & 45.62 & Vanilla-RNN & 49.57 & 47.86 & 42.10 & 44.79 \\
LSTM & 53.68 & 67.82 & 51.65 & 58.64 & 50.79 & 59.23 & 50.32 & 54.41 & LSTM & 51.28 & 57.26 & 44.07 & 49.80 \\
GRU & 54.54 & 71.30 & 53.10 & 60.87 & 52.06 & 59.91 & 49.41 & 54.15 & GRU & 51.70 & 50.42 & 45.00 & 47.55 \\
GCN & 77.85 & 78.79 & 70.02 & 74.15 & 74.21 & 75.97 & 68.35 & 71.96 & GCN & 64.01 & 63.04 & 59.96 & 61.46 \\	
DR-GCN & 81.47 & 80.89 & 72.36 & 76.39 & 78.68 & 78.91 & 71.29 & 74.91 & DR-GCN & 68.34 & 67.82 & 64.89 & 66.32 \\
TMP & 84.48 & 82.63 & 74.06 & 78.11 & 83.45 & 83.82 & 75.05 & 79.19 & TMP & 74.61 & 74.32 & 73.89 & 74.10 \\
\hline	
\textbf{AME} & \textbf{90.19} & \textbf{89.69} & \textbf{86.25} & \textbf{87.94} & \textbf{86.52} &  \textbf{86.23} & \textbf{82.07} & \textbf{84.10} &\textbf{AME} &  \textbf{80.32} & \textbf{79.08} & \textbf{78.69} & \textbf{78.88} \\
\hline
\end{tabular}
}
\caption{Performance comparison. A total of sixteen methods are investigated in the comparisons. ‘--’ denotes not applicable.}
\label{Performance_comparison}
\end{table*}

\section{Experiments}
In this section, we empirically evaluate our proposed method on two benchmark datasets, namely \emph{Ethereum} smart contract dataset (ESC) and \emph{VNT Chain} smart contract dataset (VSC). We seek to answer the following research questions:
\begin{itemize}[noitemsep,wide=0pt, leftmargin=\dimexpr\labelwidth + 2\labelsep\relax]
\item \textbf{RQ1}: Can the proposed method effectively detect the three types of vulnerabilities? How is its performance against state-of-the-art tools and neural network-based methods?
\item \textbf{RQ2}: Is the method able to provide interpretability in vulnerability detection? Can we obtain new insights from it?
\item \textbf{RQ3}: How do the different components affect the performance of the proposed approach?
\end{itemize}
Next, we first present the experimental settings, followed by answering the above research questions one by one.

\subsection{Experimental Settings} 

\paragraph{Datasets.} {(1)} ESC dataset contains 307,396 functions from 40,932 \emph{Ethereum} smart contracts. In the dataset, 5,013 functions contain at least one invocation to \emph{call.value}, making them \emph{potentially} affected by reentrancy vulnerability. 4,833 functions have the \emph{block.timestamp} statement that may cause timestamp dependence vulnerability. {(2)} VSC dataset consists of  13,761 functions from 4,170 \emph{VNT Chain} smart contracts. Around 2,925 functions have loop statements. 

\paragraph{Implementation details.} All experiments are conducted on a computer equipped with an Intel Core i7 CPU at 3.7GHz, a GPU at 1080Ti, and 32GB Memory. The expert pattern extraction tool and the graph construction tool are implemented with Python, while the neural networks are implemented with TensorFlow. Following prior works, we randomly select $80\%$ of the functions as the training set and the other $20\%$ as the test set for each dataset. The encoders in Fig.~\ref{fig_network} are implemented using three fully connected layers. The hidden state sizes of the self attention and encoder layers are 200 and 100, respectively.

\subsection{Performance Comparison (RQ1)} 
In this section, we compare our proposed approach against state-of-the-art tools and available neural network-based methods. Following existing work \cite{ijcai20}, we conduct experiments for the reentrancy and timestamp dependence vulnerability on ESC, and evaluate the infinite loop vulnerability on VSC. Metrics \emph{accuracy, recall, precision}, and \emph{F1-score} are all engaged in the comparisons. 

\subsubsection{Comparison with Conventional Detection Tools} 
We first compare our method AME with existing smart contract vulnerability detection tools including {Smartcheck} \cite{Smartcheck}, {Oyente} \cite{oyente}, {Mythril} \cite{Mythril}, {Securify} \cite{tsankov2018securify}, and {Slither} \cite{feist2019slither}. Quantitative results are summarized in Table~\ref{Performance_comparison}.

As shown in the left of Table~\ref{Performance_comparison}, we observe that: 1) conventional tools have not yet achieved a satisfactory accuracy on the reentrancy vulnerability detection. In particular, state-of-the-art tools \emph{Securify} and \emph{Slither} only achieve 71.89\% and 77.12\% accuracies. 2) Our method significantly outperforms the existing tools in reentrancy vulnerability detection. More specifically, AME achieves a 90.19\% accuracy, gaining a 13.07\% accuracy improvement over the state-of-the-art tool. Empirical evidences clearly reveal the effectiveness of our method.

Next, we evaluate all the methods on timestamp dependence vulnerability. The comparison results are shown in the middle of Table~\ref{Performance_comparison}. State-of-the-art tool \emph{Slither} obtains a 74.20\% accuracy, which is quite low. This may stem from the fact that most conventional tools handle the timestamp dependence vulnerability by blindly checking whether there is a \emph{block.timestamp} statement in the function, while ignoring whether the timestamp can truly affect a critical operation. Further, it is worth pointing out that AME keeps delivering the best performance in terms of all the four metrics. Significantly, AME gains a 12.32\% accuracy improvement over the state-of-the-art tool.

We further evaluate our method on the infinite loop vulnerability. Specifically, we compare our methods against existing infinite loop detection methods including Jolt \cite{Jolt}, SMT \cite{Smt}, PDA \cite{Pda}, and Looper \cite{Looper}. Quantitative results are illustrated in the right of Table~\ref{Performance_comparison}. We observe that AME consistently outperforms other methods by a large margin. 

By looking into the implementations of classical tools, we found that: 1) they heavily rely on a few fixed expert rules to detect vulnerabilities, e.g., \emph{Smartcheck} checks whether there exists an invocation to \emph{call.value} to detect reentrancy, and 2) the rich code semantic information and key variables in the code are not well characterized in the methods. In this respect, our work has an edge in explicitly modeling key variables and abling to handle complex semantics.

\subsubsection{Comparison with Deep Learning Methods} 
We also compare our method with available deep learning-based methods, namely Vanilla-RNN, LSTM, GRU, GCN, DR-GCN, and TMP \cite{ijcai20}. For a feasible comparison, Vanilla-RNN, LSTM, and GRU are fed with the function code sequence vectors, while GCN, DR-GCN and TMP are presented with the graph feature vectors. 

We illustrate the performance of different methods in Table~\ref{Performance_comparison}.  Results show that sequential models Vanilla-RNN, LSTM, and GRU have a relatively poor performance, while graph neural network models GCN, DR-GCN, and TMP significantly outperform them. This reconfirms that blindly treat the source code as a sequence is not suitable for vulnerability detection, while characterizing the code as a graph and employing graph neural networks is effective. Notably, AME consistently outperforms GCN, DR-GCN, and TMP by a large margin across three vulnerabilities. The  empirical evidences reveal that encoding expert patterns in networks indeed contributes to a significant performance gain.

\begin{figure}[t]
 \centering
 \subfigure{
  \includegraphics[width=0.4745\linewidth]{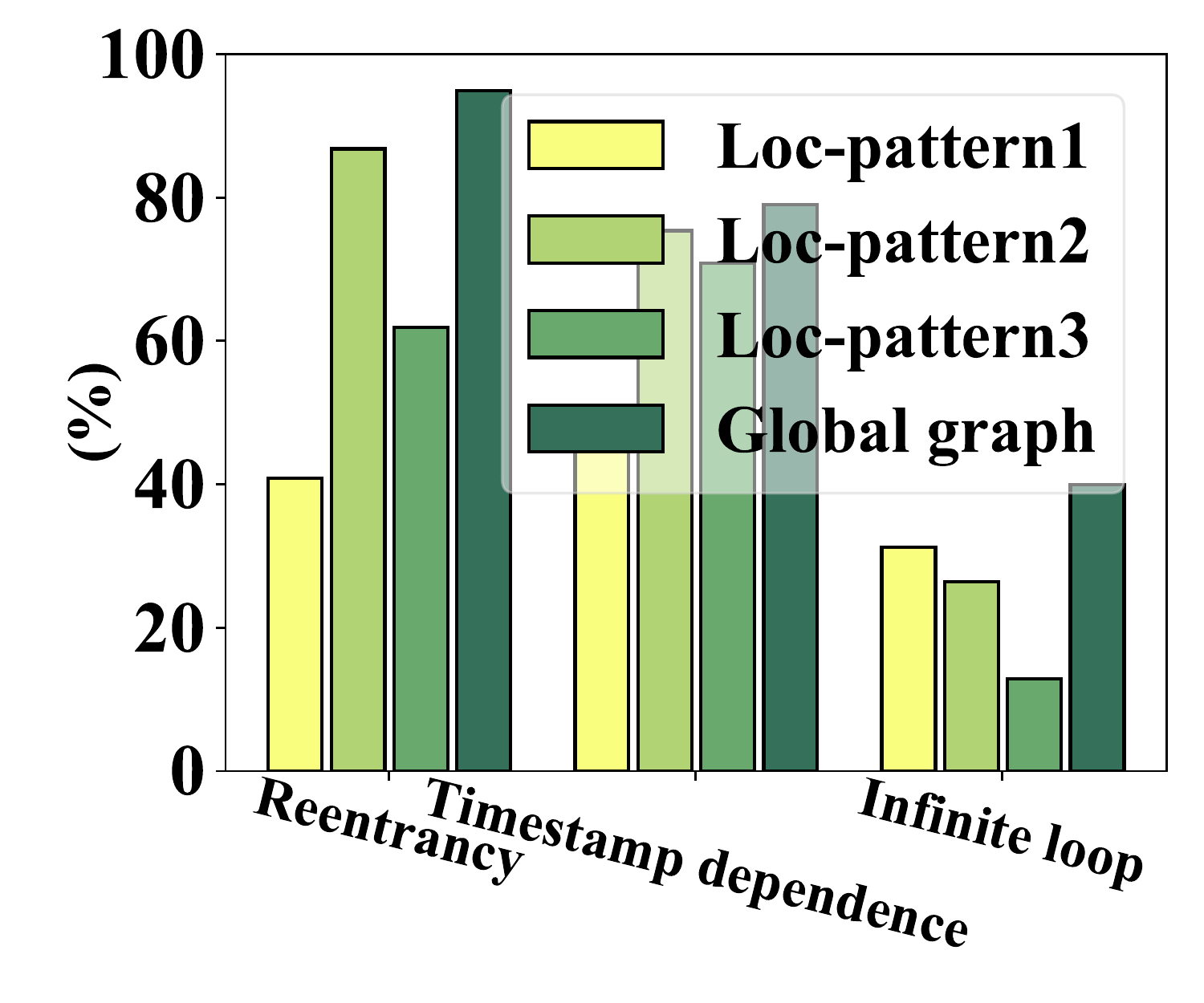}
 }
 \subfigure{
  \includegraphics[width=0.4745\linewidth]{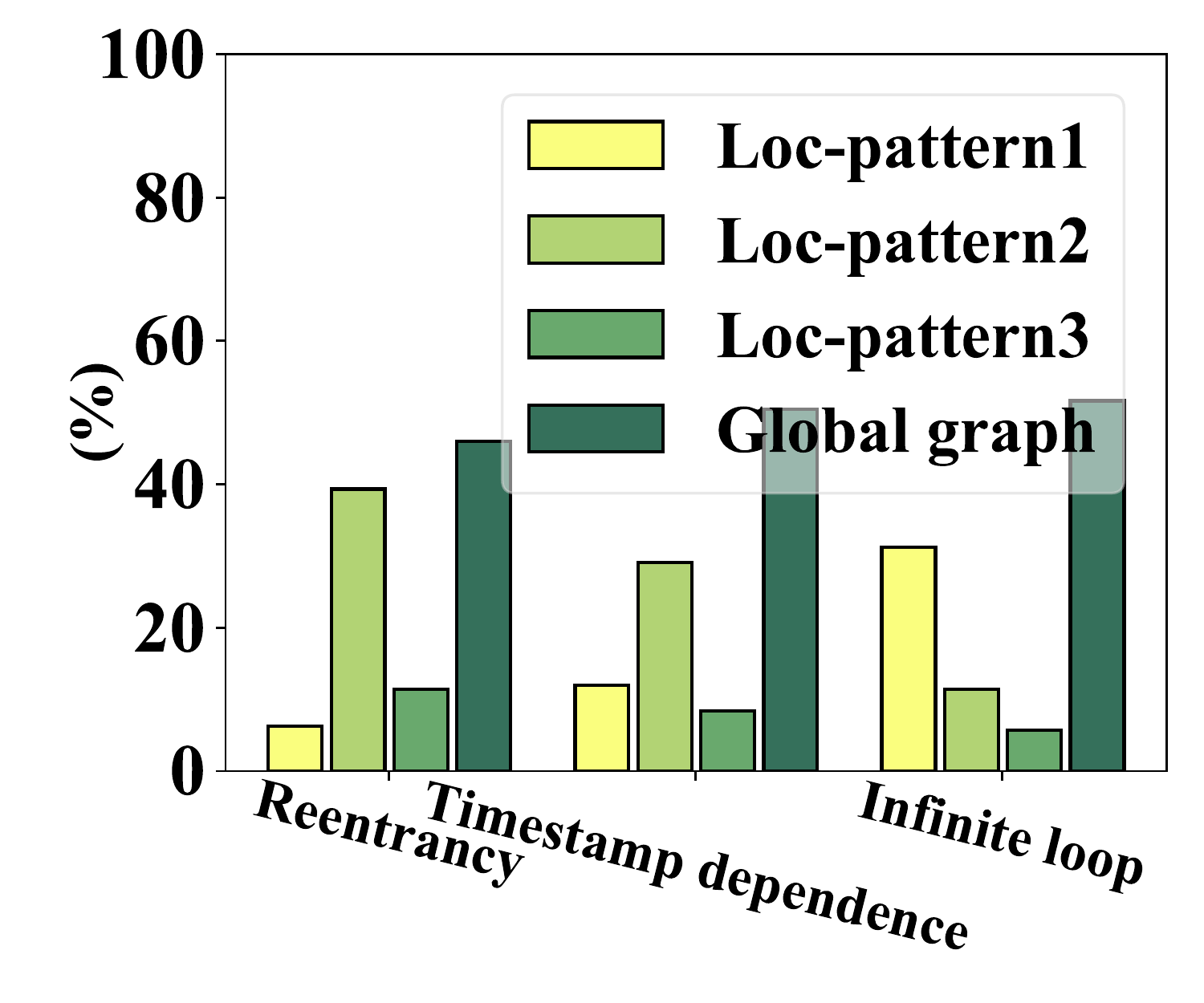}
 }
 \caption{Weight statistics of different features. Left: ratio of having a weight higher than $\sigma$. Right: ratio of having the maximum weight.}
 \label{weights_comparison}
\end{figure}

\begin{figure}[t]
 \centering
 \subfigure[Reentrancy]{
  \includegraphics[width=0.4745\linewidth]{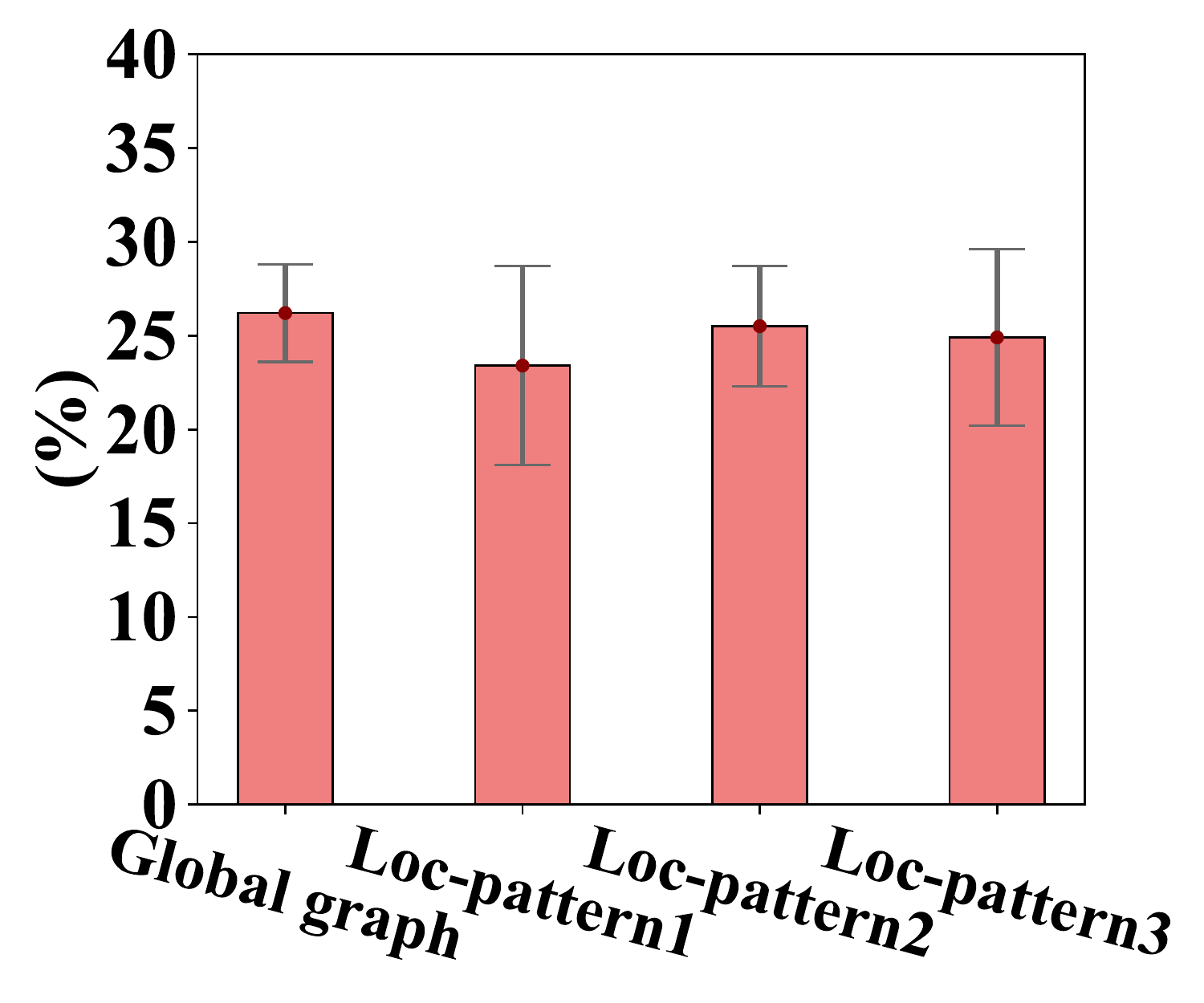}
 }
 \subfigure[Timestamp]{
  \includegraphics[width=0.4745\linewidth]{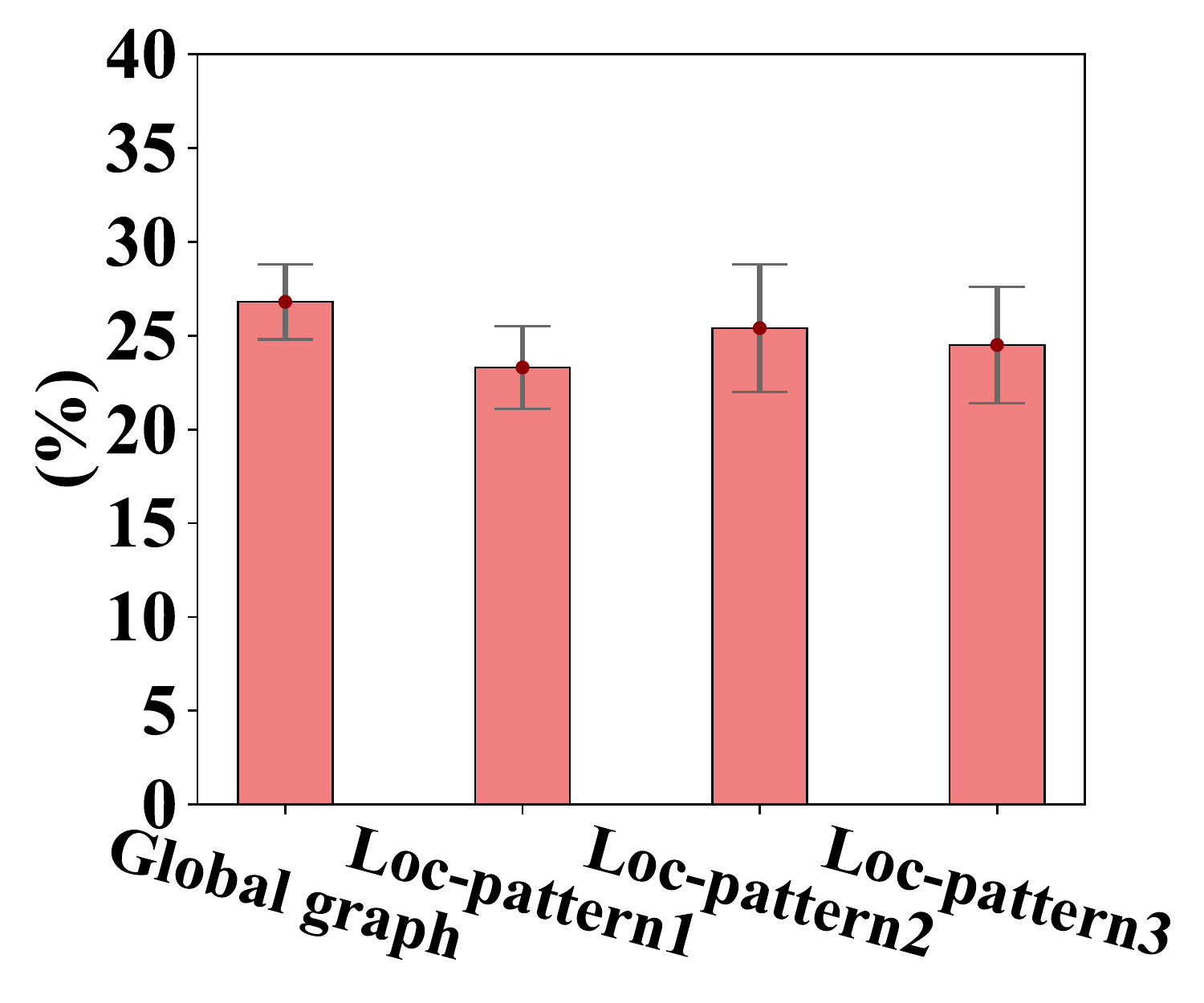}
 }
 \subfigure[Infinite loop]{
 \includegraphics[width=0.4745\linewidth]{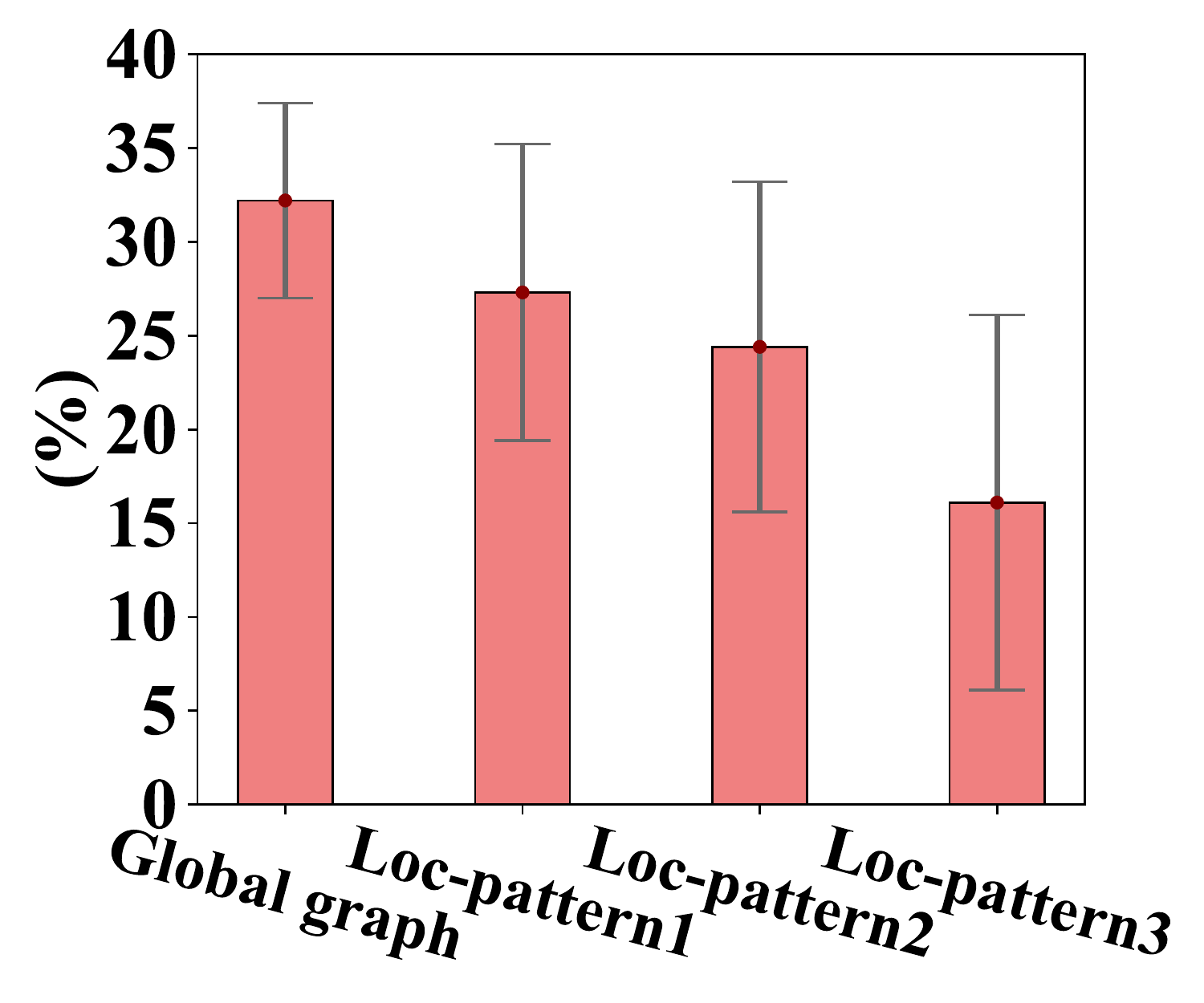}
 }
 \caption{Mean and deviation of the weights of different features.}
 \label{average_attention_weights}
\end{figure}

\subsection{Interpretability Evaluation (RQ2)} 
We now study the interpretability of the proposed AME network, and present novel insights on weight distributions of different features in vulnerability detection. We also present a case study to facilitate understanding. 

As presented in subsection~\ref{pattern_extraction}, we defined three local patterns for each vulnerability. We denote the three vulnerability-specific patterns as Loc-pattern1, Loc-pattern2, and Loc-pattern3, respectively. For example, Loc-pattern1 is \emph{enoughBalance} for reentrancy vulnerability and is \emph{timestampInvocation} for timestamp dependence vulnerability. The three patterns and the global graph feature are used to predict whether the function has the specific vulnerability.

\begin{figure*}
\centering
\includegraphics[width=16cm]{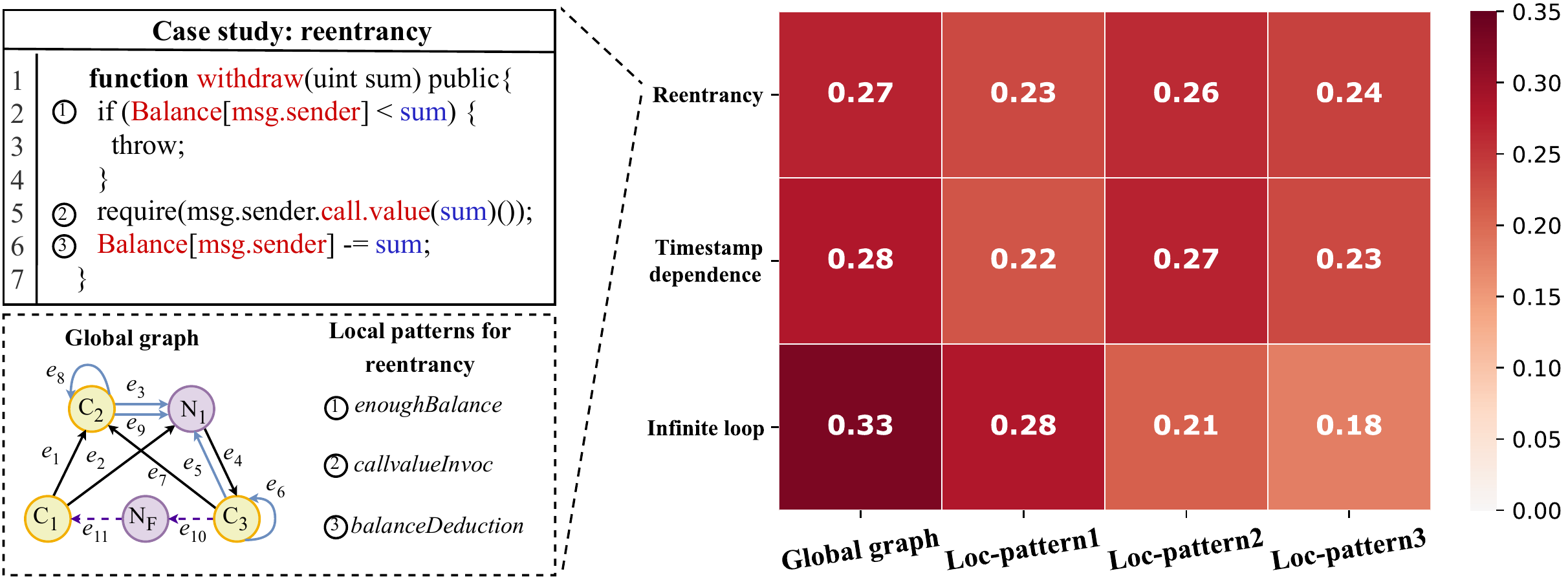}
\caption{Case study on the interpretability of our method.}
\label{case_study}
\end{figure*}

Interestingly, to figure out which features contribute the most to the detection of a specific vulnerability, we record the number that a feature possesses a weight (obtained by Eq.~7 or Eq.~8) greater than a preset threshold $\sigma=0.25$ over all tested functions. The statistics are visualized in the left of Fig.~\ref{weights_comparison}. We observe that the global graph feature usually possesses a high weight, revealing its leading role in the label prediction. Moreover, in reentrancy vulnerability detection, Loc-pattern2 (\emph{callValueInvocation}) and Loc-pattern3 (\emph{balanceDeduction}) rank second and third in the possibility of having a high weight, while Loc-pattern2 (\emph{timestampAssign}) and Loc-pattern3 (\emph{timestampContaminate}) often have high weights in timestamp dependence detection.  It is worth pointing out that once a function is predicted to have a specific vulnerability by our method, besides telling the user how significant (weight) each feature contributes to the prediction, we can also warn him/her of the expert patterns that get high weights to help find bugs. The statistics over a large number of functions also allow developers build a grand picture of the whole system.

We also compute the number that a feature has the maximum weight among all features over all tested functions. The visualized results are demonstrated in the right of Fig.~\ref{weights_comparison}. From the histogram, we notice that the global graph feature obtains the maximum weight most times for the three vulnerabilities. For each feature, we also illustrate the mean and standard deviation of its weight in Fig.~\ref{average_attention_weights}. We see that each local expert pattern possesses a considerable weight, but the average attention weight of the global graph feature is the highest. This may come from the fact that the global graph conveys the control- and data- dependencies of a specific vulnerability and contains richer global semantic information than local expert patterns.

\paragraph{Case study.} We further present a case study in Fig.~\ref{case_study}, where the $withdraw$ function is a real-world smart contract function that has a reentrancy vulnerability. We analyzed the code to detect whether it has the three vulnerabilities. The left of Fig.~\ref{case_study} illustrates the global graph and the three local patterns of reentrancy. The right of Fig.~\ref{case_study} shows the interpretable weights of the graph feature and each local pattern feature. For example, the method detects the function has the reentrancy vulnerability. To make the detection decision, the weight of the global graph feature is 0.27, the weights of Loc-pattern1 (\emph{enoughBalance}),  Loc-pattern2 (\emph{callValueInvocation}), and Loc-pattern3 (\emph{balanceDeduction}) are 0.23, 0.26, and 0.24, respectively. Thus, our system is able to clearly explain the reasons behind the predictions.

\begin{figure}[t]
 \centering
 \subfigure[Removing graph feature]{
  \includegraphics[width=0.4745\linewidth]{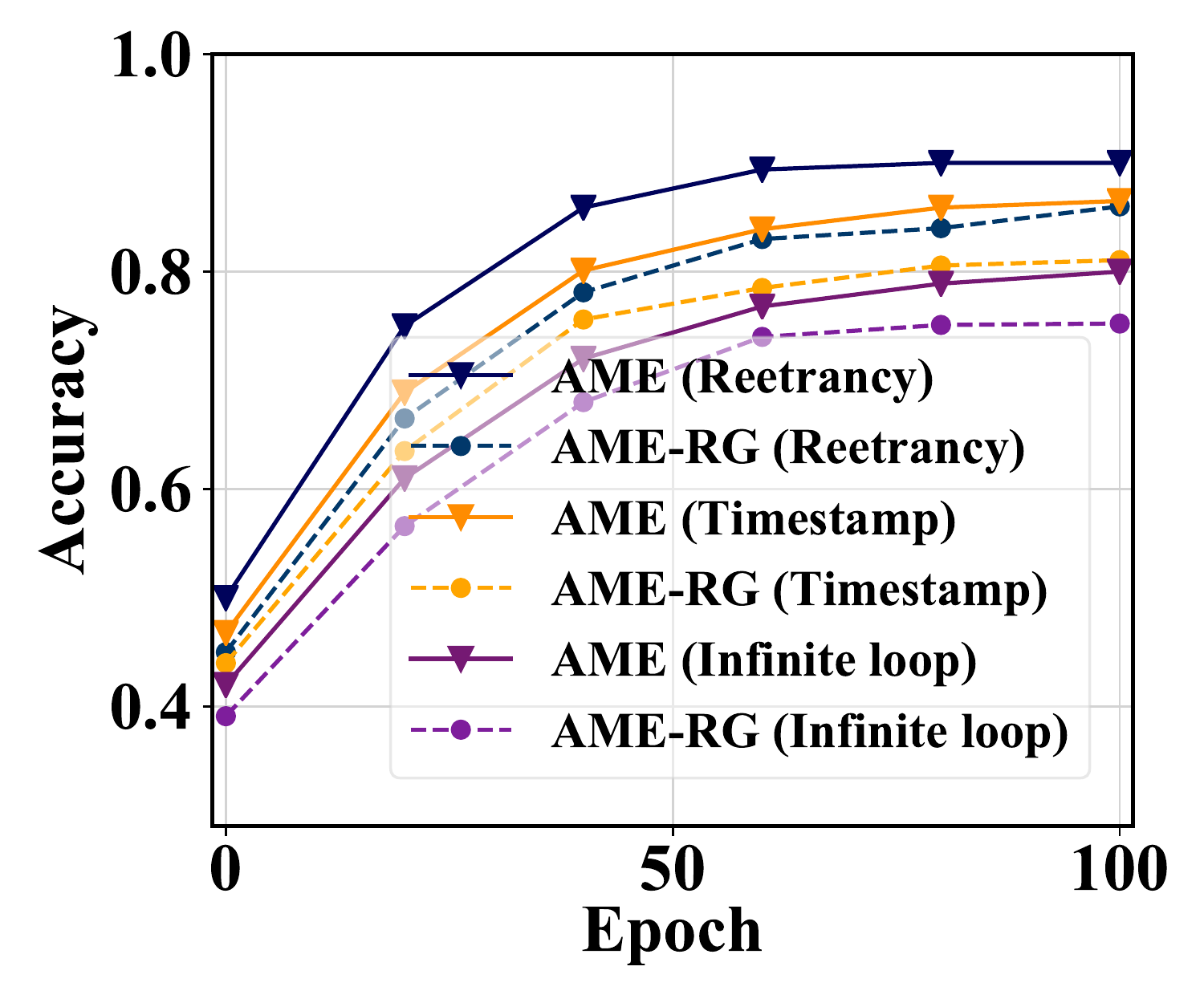}
 }
 \subfigure[Removing patterns]{
  \includegraphics[width=0.4745\linewidth]{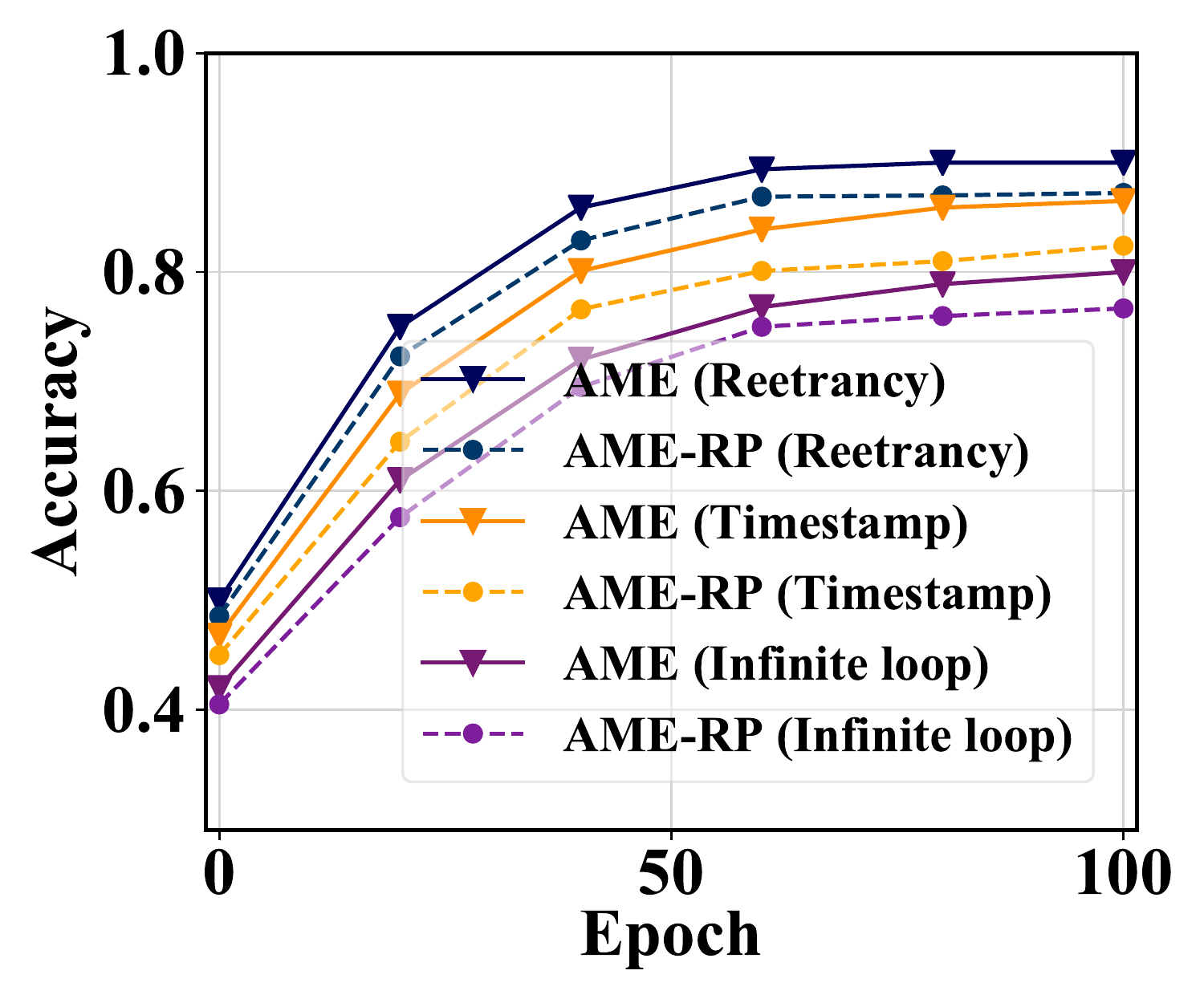}
 }

 \caption{Effect of removing modules on three vulnerabilities.}

 \label{feature_combination}
\end{figure}

\subsection{Effects of Removing Expert Patterns and Graph Feature (RQ3)} 
By default, we combine the \emph{graph feature} with \emph{expert patterns} for vulnerability detection. We are interested in exploring the effect of removing them respectively.  To this aim, we first remove the code semantic graph construction module, and only use the expert pattern features as the inputs. We denote this variant as AME-RG (RG represents removing graph feature). We also try removing the expert pattern extraction module and using the graph feature only, which is denoted as AME-RP (RP represents removing patterns, which is almost identical to the TMP). Fig.~\ref{feature_combination} demonstrates the comparison results on three vulnerabilities, where the solid curves demonstrate the accuracy of AME over different epochs, and the dashed curves show its variants. Different colors represent different vulnerabilities. Clearly, the performance of AME is consistently better compared to its variants across all epochs, revealing that combining local expert patterns with global graph features is necessary and important to improve the detection performance.  Moreover, removing the neural network extracted graph feature exhibits a higher performance drop than removing the expert patterns.

\section{Related Work}
Early work on smart contract vulnerability detection verifies smart contracts by employing formal methods \cite{bhargavan2016formal,Hirai,Grishchenko}. For example, \cite{bhargavan2016formal} introduces a framework, translating \emph{Solidity} code and the EVM (Ethereum Virtual Machine)  bytecode into the input of an existing verification system. \cite{Hirai} proposes a formal model for EVM (Ethereum Virtual Machine) and reasons smart contracts using the Isabelle/HOL tool. Another stream of work relies on symbolic analysis and dynamic execution. Oyente  performs symbolic execution on contract functions and flags bugs based on simple patterns. Zeus \cite{zeus} leverages abstract interpretation and symbolic model checking. \cite{tsankov2018securify} introduces compliance (negative) and violation (positive) patterns to filter false warnings. \cite{contractfuzzer} presents ContractFuzzer to identify vulnerabilities by fuzzing and runtime behavior monitoring during execution. Sereum \cite{sereum} uses taint analysis to monitor runtime data flows during smart contract execution for vulnerability detection. Recently, a few attempts have been made to study using deep neural networks. \cite{peng} constructs the sequential \emph{contract snippet} and feeds them into the BLSTM-ATT model. \cite{ijcai20} proposes to convert the source code of contract into the \emph{contract graph} and constructs graph neural networks as the detection model. \cite{wang2020contractward} proposes extracting bigram features from operation codes of smart contracts. \cite{tkde2021} proposes to combine expert rules with neural networks for improving the detection accuracy. However, \cite{tkde2021} suffers from poor explainability and fail to investigate the significance of different features.

\section{Conclusion}
In this paper, we explore combining deep learning with classical expert patterns in an explainable way for smart contract vulnerability detection. This system consists of both neural networks and automatic expert pattern extraction tools. Interestingly, the model is able to obtain explainable fine-grained details and a grand picture of the weight distributions. Extensive experiments show that our method significantly outperforms state-of-the-art approaches. We believe our work is an important step towards explainable and accurate contract vulnerability detection.

\section*{Acknowledgements}
This paper is supported by the Natural Science Foundation of Zhejiang Province, China (No. LQ19F020001), the National Natural Science Foundation of China (No. 61902348), and the Key R\&D Program of Zhejiang Province (No. 2021C01104).

\bibliographystyle{named}
\bibliography{ijcai21}

\begin{thebibliography}{}

\bibitem[\protect\citeauthoryear{Bhargavan \bgroup \em et al.\egroup
  }{2016}]{bhargavan2016formal}
Karthikeyan Bhargavan, Antoine Delignat-Lavaud, C{\'e}dric Fournet, Anitha
  Gollamudi, Georges Gonthier, Nadim Kobeissi, Natalia Kulatova, Aseem Rastogi,
  Thomas Sibut-Pinote, Nikhil Swamy, et~al.
\newblock Formal verification of smart contracts: Short paper.
\newblock In {\em Proceedings of the 2016 ACM Workshop on Programming Languages
  and Analysis for Security}, pages 91--96, 2016.

\bibitem[\protect\citeauthoryear{Burnim \bgroup \em et al.\egroup
  }{2009}]{Looper}
Jacob Burnim, Nicholas Jalbert, Christos Stergiou, and Koushik Sen.
\newblock Looper: Lightweight detection of infinite loops at runtime.
\newblock In {\em Proceedings of the International Conference on Automated
  Software Engineering}, pages 161--169. IEEE Computer Society, 2009.

\bibitem[\protect\citeauthoryear{Carbin \bgroup \em et al.\egroup
  }{2011}]{Jolt}
Michael Carbin, Sasa Misailovic, Michael Kling, and Martin~C Rinard.
\newblock Detecting and escaping infinite loops with jolt.
\newblock In {\em European Conference on Object-Oriented Programming}, pages
  609--633. Springer, 2011.

\bibitem[\protect\citeauthoryear{Ding \bgroup \em et al.\egroup }{2019}]{dapps}
Yi~Ding, Jun Jin, Jinglun Zhang, Zhongyi Wu, and Kai Hu.
\newblock Sc-rbac: A smart contract based rbac model for dapps.
\newblock In {\em International Conference on Human Centered Computing}, pages
  75--85. Springer, 2019.

\bibitem[\protect\citeauthoryear{Feist \bgroup \em et al.\egroup
  }{2019}]{feist2019slither}
Josselin Feist, Gustavo Grieco, and Alex Groce.
\newblock Slither: a static analysis framework for smart contracts.
\newblock In {\em 2019 IEEE/ACM 2nd International Workshop on Emerging Trends
  in Software Engineering for Blockchain (WETSEB)}, pages 8--15. IEEE, 2019.

\bibitem[\protect\citeauthoryear{Grishchenko \bgroup \em et al.\egroup
  }{2018}]{Grishchenko}
Ilya Grishchenko, Matteo Maffei, and Clara Schneidewind.
\newblock A semantic framework for the security analysis of ethereum smart
  contracts.
\newblock In {\em International Conference on Principles of Security and
  Trust}, pages 243--269, 2018.

\bibitem[\protect\citeauthoryear{Hewa \bgroup \em et al.\egroup
  }{2020}]{hewa2020survey}
Tharaka Hewa, Mika Ylianttila, and Madhusanka Liyanage.
\newblock Survey on blockchain based smart contracts: Applications,
  opportunities and challenges.
\newblock {\em Journal of Network and Computer Applications}, page 102857,
  2020.

\bibitem[\protect\citeauthoryear{Hirai}{2017}]{Hirai}
Yoichi Hirai.
\newblock Defining the ethereum virtual machine for interactive theorem
  provers.
\newblock In {\em International Conference on Financial Cryptography and Data
  Security}, pages 520--535, 2017.

\bibitem[\protect\citeauthoryear{Ibing and Mai}{2015}]{Pda}
Andreas Ibing and Alexandra Mai.
\newblock A fixed-point algorithm for automated static detection of infinite
  loops.
\newblock In {\em International Symposium on High Assurance Systems
  Engineering}, pages 44--51. IEEE, 2015.

\bibitem[\protect\citeauthoryear{Jiang \bgroup \em et al.\egroup
  }{2018}]{contractfuzzer}
Bo~Jiang, Ye~Liu, and WK~Chan.
\newblock Contractfuzzer: Fuzzing smart contracts for vulnerability detection.
\newblock In {\em ASE}, pages 259--269. IEEE, 2018.

\bibitem[\protect\citeauthoryear{Kalra \bgroup \em et al.\egroup }{2018}]{zeus}
Sukrit Kalra, Seep Goel, Mohan Dhawan, and Subodh Sharma.
\newblock Zeus: Analyzing safety of smart contracts.
\newblock In {\em {NDSS}}, 2018.

\bibitem[\protect\citeauthoryear{Kling \bgroup \em et al.\egroup }{2012}]{Smt}
Michael Kling, Sasa Misailovic, Michael Carbin, and Martin Rinard.
\newblock Bolt: on-demand infinite loop escape in unmodified binaries.
\newblock {\em ACM SIGPLAN Notices}, 47(10):431--450, 2012.

\bibitem[\protect\citeauthoryear{Liu \bgroup \em et al.\egroup
  }{2021}]{tkde2021}
Zhenguang Liu, Peng Qian, Xiaoyang Wang, Yuan Zhuang, Lin Qiu, and Xun Wang.
\newblock Combining graph neural networks with expert knowledge for smart
  contract vulnerability detection.
\newblock {\em Transactions on Knowledge and Data Engineering}, page Accepted,
  2021.

\bibitem[\protect\citeauthoryear{Lorenz \bgroup \em et al.\egroup
  }{2018}]{Multisig}
Breidenbach Lorenz, Daian Phil, Juels Ari, and Sirer Emin.
\newblock An in-depth look at the parity multisig bug.
\newblock
  \url{https://hackingdistributed.com/2017/07/22/deep-dive-parity-bug/}, 2018.
\newblock Accessed: 2021-01-10.

\bibitem[\protect\citeauthoryear{Luu \bgroup \em et al.\egroup }{2016}]{oyente}
Loi Luu, Duc-Hiep Chu, Hrishi Olickel, Prateek Saxena, and Aquinas Hobor.
\newblock Making smart contracts smarter.
\newblock In {\em Proceedings of the 2016 ACM SIGSAC Conference on Computer and
  Communications Security}, pages 254--269, 2016.

\bibitem[\protect\citeauthoryear{Mueller}{2017}]{Mythril}
Bernhard Mueller.
\newblock A framework for bug hunting on the ethereum blockchain.
\newblock \url{https://github.com/ConsenSys/mythril}, 2017.
\newblock Accessed: 2021-01-10.

\bibitem[\protect\citeauthoryear{{Qian} \bgroup \em et al.\egroup
  }{2020}]{peng}
P.~{Qian}, Z.~{Liu}, Q.~{He}, R.~{Zimmermann}, and X.~{Wang}.
\newblock Towards automated reentrancy detection for smart contracts based on
  sequential models.
\newblock {\em IEEE Access}, 8:19685--19695, 2020.

\bibitem[\protect\citeauthoryear{Rodler \bgroup \em et al.\egroup
  }{2019}]{sereum}
Michael Rodler, Wenting Li, Ghassan~O. Karame, and Lucas Davi.
\newblock Sereum: Protecting existing smart contracts against re-entrancy
  attacks.
\newblock In {\em Proceedings of the {NDSS}}, 2019.

\bibitem[\protect\citeauthoryear{Tann \bgroup \em et al.\egroup
  }{2018}]{Wesley}
Wesley~Joon{-}Wie Tann, Xing~Jie Han, Sourav~Sen Gupta, and Yew{-}Soon Ong.
\newblock Towards safer smart contracts: {A} sequence learning approach to
  detecting vulnerabilities.
\newblock {\em CoRR}, 2018.

\bibitem[\protect\citeauthoryear{Tikhomirov \bgroup \em et al.\egroup
  }{2018}]{Smartcheck}
Sergei Tikhomirov, Ekaterina Voskresenskaya, Ivan Ivanitskiy, Ramil Takhaviev,
  Evgeny Marchenko, and Yaroslav Alexandrov.
\newblock Smartcheck: Static analysis of ethereum smart contracts.
\newblock In {\em International Workshop on Emerging Trends in Software
  Engineering for Blockchain}, pages 9--16. IEEE, 2018.

\bibitem[\protect\citeauthoryear{Tsankov \bgroup \em et al.\egroup
  }{2018}]{tsankov2018securify}
Petar Tsankov, Andrei Dan, Dana Drachsler-Cohen, Arthur Gervais, Florian
  Buenzli, and Martin Vechev.
\newblock Securify: Practical security analysis of smart contracts.
\newblock In {\em Proceedings of the 2018 ACM SIGSAC Conference on Computer and
  Communications Security}, pages 67--82, 2018.

\bibitem[\protect\citeauthoryear{Wang \bgroup \em et al.\egroup
  }{2019}]{wang2019blockchain}
Shuai Wang, Liwei Ouyang, Yong Yuan, Xiaochun Ni, Xuan Han, and Fei-Yue Wang.
\newblock Blockchain-enabled smart contracts: architecture, applications, and
  future trends.
\newblock {\em IEEE Transactions on Systems, Man, and Cybernetics: Systems},
  49(11):2266--2277, 2019.

\bibitem[\protect\citeauthoryear{Wang \bgroup \em et al.\egroup
  }{2020}]{wang2020contractward}
Wei Wang, Jingjing Song, Guangquan Xu, Yidong Li, Hao Wang, and Chunhua Su.
\newblock Contractward: Automated vulnerability detection models for ethereum
  smart contracts.
\newblock {\em IEEE Transactions on Network Science and Engineering}, 2020.

\bibitem[\protect\citeauthoryear{Zhuang \bgroup \em et al.\egroup
  }{2020}]{ijcai20}
Yuan Zhuang, Zhenguang Liu, Peng Qian, Qi~Liu, Xiang Wang, and Qinming He.
\newblock Smart contract vulnerability detection using graph neural network.
\newblock In {\em {IJCAI}}, pages 3283--3290, 2020.

\end{thebibliography}

\end{document}